\pdfoutput=1

\documentclass[11pt]{article}

\usepackage[]{EMNLP2022}

\usepackage{comment}
\usepackage{times}
\usepackage{latexsym}
\usepackage{amssymb}
\usepackage{multirow}
\usepackage{xspace}
\usepackage{booktabs}
\usepackage{hyperref}
\usepackage{todonotes}
\usepackage{arydshln}
\usepackage[T1]{fontenc}
\usepackage[utf8]{inputenc}
\usepackage{CJKutf8}
\usepackage{microtype}
\usepackage{xcolor}
\usepackage{graphicx}
\usepackage{caption}
\usepackage{subcaption}

\usepackage{array}

\definecolor{maroon}{RGB}{191, 96, 96}

\definecolor{purple}{RGB}{100, 0, 200}

\definecolor{mint}{rgb}{0.24, 0.71, 0.54}

\newcolumntype{L}[1]{>{\raggedright\let\newline\\\arraybackslash\hspace{0pt}}m{#1}}
\newcolumntype{C}[1]{>{\centering\let\newline\\\arraybackslash\hspace{0pt}}m{#1}}
\newcolumntype{R}[1]{>{\raggedleft\let\newline\\\arraybackslash\hspace{0pt}}m{#1}}

\makeatletter
\def\adl@drawiv#1#2#3{%
        \hskip.5\tabcolsep
        \xleaders#3{#2.5\@tempdimb #1{1}#2.5\@tempdimb}%
                #2\z@ plus1fil minus1fil\relax
        \hskip.5\tabcolsep}
\newcommand{\cdashlinelr}[1]{%
  \noalign{\vskip\aboverulesep
           \global\let\@dashdrawstore\adl@draw
           \global\let\adl@draw\adl@drawiv}
  \cdashline{#1}
  \noalign{\global\let\adl@draw\@dashdrawstore
           \vskip\belowrulesep}}
\makeatother

\newcommand{\ie}{i.\,e. }

\newcommand{\datasetname}{\textsc{HumSet}\xspace}

\title{\datasetname: Dataset of Multilingual Information Extraction and Classification for Humanitarian Crisis Response}

\author{Selim Fekih$^1$~~~~Nicolò Tamagnone$^1$~~~~Benjamin Minixhofer$^3$~~~~Ranjan Shrestha$^2$\\
{\bf Ximena Contla}$^1$~~~~{\bf Ewan Oglethorpe}$^1$~~~~{\bf Navid Rekabsaz}$^3$  \\
  $^1$Data Friendly Space~~~~$^2$ToggleCorp Solutions\\
  $^3$Johannes Kepler University Linz, LIT AI Lab, Austria \\
  \texttt{\{selim, nico, ximena, ewan\}@datafriendlyspace.org}\\
  \texttt{ranjan.shrestha@togglecorp.com}\\
  \texttt{\{benjamin.minixhofer, navid.rekabsaz\}@jku.at}
}

\begin{document}
\maketitle
\begin{abstract}
Timely and effective response to humanitarian crises requires quick and accurate analysis of large amounts of text data -- a process that can highly benefit from expert-assisted NLP systems trained on validated and annotated data in the humanitarian response domain. To enable creation of such NLP systems, we introduce and release \datasetname, a novel and rich multilingual dataset of humanitarian response documents annotated by experts in the humanitarian response community. The dataset provides documents in three languages (English, French, Spanish) and covers a variety of humanitarian crises from 2018 to 2021 across the globe. For each document, \datasetname provides selected snippets (entries) as well as assigned classes to each entry annotated using common humanitarian information analysis frameworks. \datasetname also provides novel and challenging entry extraction and multi-label entry classification tasks. In this paper, we take a first step towards approaching these tasks and conduct a set of experiments on  Pre-trained Language Models (PLM) to establish strong baselines for future research in this domain. The dataset is available at \url{https://blog.thedeep.io/humset/}.
\end{abstract}

\section{Introduction}
\label{sec:introduction}

During humanitarian crises caused by reasons ranging from natural disasters, wars ,or epidemics such as COVID-19, a timely and effective humanitarian response highly depends on fast and accurate analysis of relevant data to yield key information. Early in the response phase, namely in the first 72 hours after a disaster strikes, the humanitarian response analysts in international organizations\footnote{Such as the International Federation of Red Cross (IFRC), the United Nations High Commissioner for Refugees (UNHCR), or the United Nations Office for the Coordination of Humanitarian Affairs (UNOCHA)} review large amounts of data loosely or strongly relevant to the crisis to gain situational awareness. A large portion of this data appears in the form of secondary data sources \ie reports, news, and other forms of text data, and is integral in revealing which type of relief activities to undertake. Analysis in this phase involves extracting key information and organizing it according to sets of pre-defined domain-specific structures and guidelines, referred to as \emph{humanitarian analysis frameworks}.

While typically only small workforces are available to analyze such information, an automatic document processing system can significantly help analysts save time in the overall humanitarian response cycle. To facilitate such systems, we introduce and release \datasetname, a unique and rich dataset of document analysis in the humanitarian response domain. \datasetname is curated by humanitarian analysts and covers various disasters around the globe that occurred from 2018 to 2021 in 46 humanitarian response projects. The dataset consists of approximately 17K annotated documents in three languages of English, French, and Spanish, originally taken from publicly-available resources.\footnote{\url{https://app.thedeep.io/terms-and-privacy/}} For each document, analysts have identified informative snippets (entries) with respect to common humanitarian frameworks and assigned one or many classes to each entry (details in \S\ref{sec:dataset}).

\datasetname provides a large dataset for the training and evaluation of entry extraction and classification models, enabling the research and development of further NLP systems in the humanitarian response domain. We take the first step in this direction, by studying the performance of a set of strong baseline models (details in \S\ref{sec:experiments}). Our released dataset expands the previously provided collection by \citet{yela-bello-etal-2021-multihumes} with a more recent and comprehensive set of projects, as well as additional classification labels. Other similar datasets in the humanitarian domain, \citet{imran2016twitter} present human-annotated Twitter corpora collected during 19 different crises between 2013 and 2015, \citet{alam2021crisisbench} provide a combination of various social-media crisis-related existing datasets, and \citet{9137480} and later \citet{alharbi-lee-2021-kawarith} publish Arabic Twitter classification datasets for crisis events. \datasetname, in contrast to the current resources which mostly originated from social media, is created by humanitarian experts through an annotation process on official documents and news from the most recognized humanitarian agencies, conferring high reliability, continuous updating, and accurate geolocation information. 

\section{\datasetname Dataset}
\label{sec:dataset}


The collection originated from a multi-organizational platform called \emph{the Data Entry and Exploration Platform} (DEEP),\footnote{\url{https://thedeep.io/}} developed and maintained by Data Friendly Space (DFS)\footnote{Data Friendly Space is a U.S. based international non-governmental organization (INGO) \url{https://datafriendlyspace.org/}} The platform facilitates classifying primarily qualitative information with respect to analysis frameworks and allows for collaborative classification and annotation of secondary data. The dataset is available at \url{https://blog.thedeep.io/humset/}. 

\subsection{Dataset Overview} \label{subsection:datasetoverview}
\datasetname consists of data used to inform 46 humanitarian response operations across the globe. 24 responses were in Central/South America, 14 in Africa, and 8 in Asia (detailed countries can be found in Table~\ref{table:projs-per-region} in Appendix). 

For each project, documents, referred to as \emph{leads}, related to a particular humanitarian crisis are collected, analyzed, and annotated. The annotated documents in the dataset mostly consist of recently released information, with 79\% of the documents being released in 2020 and 2021 (Table~\ref{table:publidh-year} in Appendix), and 90\% of all documents being sourced from websites (see Table~\ref{table:web-sources} in Appendix for the most commonly used platforms).
Documents are selected from different sources, ranging from official reports by humanitarian organizations to international and national media articles. Overall, documents consist of files in PDF format ($70.4\%$) and HTML pages ($29.6\%$) with an average length of $\sim\!2$K words. The number of documents analyzed per project varies, ranging from 2 to 2,266. 

The relevant snippets of texts, referred to as \emph{entries}, in each document are annotated by humanitarian experts. The dataset provides an average of $\sim\!10$ entries per document, and an average length of $\sim\!65$ words per entry. Overall, \datasetname is composed of 148,621 tagged entries, selected from 16,857 documents, and in three languages: English (61.3\%), French (20.4\%) and Spanish (18.3\%). The list of projects as well as the number of documents and annotated entries per project is reported in Table~\ref{table:projects} in Appendix. Figure~\ref{fig:distributions} shows the distribution of the number of tagged documents per project, as well as the number of tokens per document and entry.

\begin{figure}[t]
\centering
\subcaptionbox{}{\includegraphics[width=0.4\textwidth]{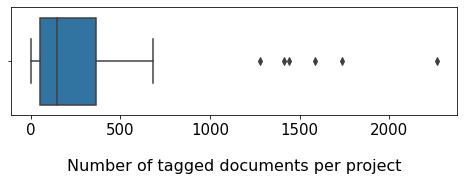}\label{fig:doc_per_project}}
\subcaptionbox{}{\includegraphics[width=0.4\textwidth]{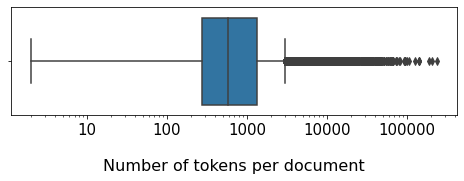}\label{fig:tokens_per_document}}
\subcaptionbox{}{\includegraphics[width=0.4\textwidth]{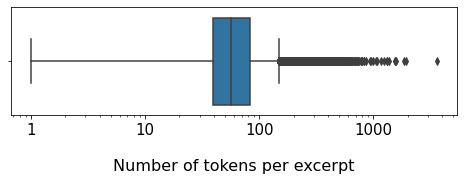}\label{fig:tokens_per_excerpt}}
\centering
\caption{(a) Distribution of documents per project. (b) Log-scale distribution of tokens\footnotemark{} per document. (c) Log-scale distribution of tokens per entry.}
\label{fig:distributions}
\vspace{-4mm}
\end{figure}

\footnotetext{Tokenized using \texttt{word\_tokenize} in NLTK v3.7 library~\cite{bird-loper-2004-nltk}.}

\begin{table}[t]
\centering
\small
\begin{tabular}{l c L{4.1cm}}
\toprule
\multirow{2}{*}{Categories} & \multicolumn{2}{c}{}\\
& \# & \multicolumn{1}{c}{Tags}\\\midrule
Sectors & 11 & Agriculture, Cross-sector, Education, Food Security, Health, 
Livelihoods, Logistics, Nutrition, Protection, Shelter, WASH (Water, Sanitation \& Hygiene)\\ \midrule
Pillars~1D & 7 & Context, COVID-19, Displacement, Humanitarian Access, Information \& Communication, Casualties, Shock/Event  \\ \midrule
Subpillars~1D & 33 & Details in Table~\ref{tbl:framework:subpillar1d} in Appendix \\ \midrule
Pillars~2D & 6 & Capacities \& Response, Humanitarian Conditions, Impact, At Risk, Priority Needs, Priority Interventions  \\ \midrule
Subpillars~2D & 18 &  Details in Table~\ref{tbl:framework:subpillar2d} in Appendix \\ 
 \bottomrule
\end{tabular}
\caption{Overview of humanitarian analysis framework.}
\label{tbl:framework:overview}
\vspace{-4mm}
\end{table}

\subsection{Humanitarian Analysis Frameworks and Data Annotation Process} \label{subsection:framework}
The concept of \emph{analytical frameworks} originated in the social sciences \cite{ragin2011constructing}, but can be considered
foundational and indispensable in numerous research
fields. An analytical framework is a set of methodologies and guidelines to facilitate data collection, collation, and analysis, helping to understand what information will be useful and what can be discarded.

In the humanitarian domain, an analytical framework (or analysis framework) not only assists decision-makers to speed up humanitarian response and disaster relief but also enables various groups to share resources~\cite{zhang2002knowledge}. When starting a response or project, humanitarian organizations create or more often use an existing analysis framework, which covers the generic but also specific needs of the work. Our data originally contained 11 different frameworks. As there are high similarities across frameworks, we created a common framework, which we refer to as \emph{humanitarian analysis framework}. This framework covers the framework dimensions of all projects. We build our custom set of tags by mapping the original tags in other frameworks to ours. More specifically, our analysis framework consists of three categories: Sectors (11 tags), Subpillars~1D (33 tags), and Subpillars~2D (18 tags). Pillars/Subpillars 1D, and 2D  have a hierarchical structure, consisting of a two-leveled tree hierarchy (Pillars to Subpillars). The list and the number of tags present for each category are reported in Table~\ref{tbl:framework:overview}.


For each project, documents relevant to understanding the situation, unmet needs, and underlying factors are captured and uploaded to the DEEP platform. From these sources, entries of text are selected and categorized into an analysis framework. Humanitarian annotators are trained in specific projects to follow analytical standards and thinking to review secondary data.

This process eventually results in annotating and organizing the data according to the humanitarian analysis framework. As the \datasetname dataset is created in a real-world scenario, the distribution of annotated entries is skewed, with 33 tags being present in less than 2\% of data. Tables~\ref{table:sectors-nb}, \ref{table:subpillars1d-nb}, and \ref{table:subpillars2d-nb} in Appendix show the detailed number and proportions of the annotated entries in Sectors, Subpillars~1D, and 2D, respectively. Figure~\ref{fig:histogram} in Appendix reports the distribution of tags in dataset.
 

\subsection{NLP Tasks} 
\paragraph{Entry Extraction Task.}
The first step for humanitarian taggers in analyzing a document is finding entries containing relevant information. A piece of text or information is considered relevant if it meaningfully contains at least one tag present in the given humanitarian analytical framework. Since documents often contain a large amount of information (Figure \ref{fig:distributions}), it is extremely beneficial to automate the process of entry identification, and this is the first task of this research. This can be seen as an extractive summarization task \ie selecting a subset of passages that contain relevant information from the given document. However, the entries do not necessarily follow the common units of text such as sentence and paragraph and can appear in various lengths. In fact, only 38.8\% of entries consist of full sentences, and the rest are snippets that are shorter or longer than sentences. This limits the direct applicability of prior approaches to extractive summarization~\cite{liu-lapata-2019-text,zhou-etal-2018-neural-document}, and makes the task particularly challenging for NLP research.


\paragraph{Multi-label Entry Classification Task.}
After selecting the most relevant entries within a document, the next step is to categorize them according to the humanitarian analysis framework (Table~\ref{tbl:framework:overview}). An automatic suggestion on which tag to choose from a large number of possibilities can be decisive in speeding up the annotation process. For each category, more than one tag can be assigned to an entry. Hence, we can view this task as multi-label classification.

\begin{table*}[t]
\centering
\begin{tabular}{l r r | r r | r r | r r | r r}
\toprule
\multirow{2}{*}{Model} & \multicolumn{2}{c}{Sectors} & \multicolumn{2}{c}{Pillars~1D} & \multicolumn{2}{c}{Subpillars~1D} & \multicolumn{2}{c}{Pillars~2D} & \multicolumn{2}{c}{Subpillars~2D} \\
 & Prec. & F1 & Prec. & F1 & Prec. & F1 & Prec. & F1 & Prec. & F1 \\ \midrule
Random Baseline &0.09&0.09  &0.06  &0.06  &0.01  &0.01  &0.13  &0.13  &0.05  &0.05 \\
\midrule
FastText & 0.71 & 0.61 & \textbf{0.56}  & 0.38  & \textbf{0.58}  & 0.33  & 0.59 & 0.45 & 0.48 & 0.33 \\
XtremeDistil$_{\mathrm{l6-h256}}$ & 0.56 & 0.58 & 0.35  & 0.36  & 0.20 & 0.20 & 0.51 & 0.55 & 0.28  & 0.29 \\
XLM-R$_{\mathrm{Base}}$ & \textbf{0.71} & \textbf{0.73} & 0.49  & \textbf{0.53} & 0.45  & \textbf{0.38} & \textbf{0.63}  & \textbf{0.63}  & \textbf{0.51}  & \textbf{0.40} \\
 \bottomrule
\end{tabular}
\caption{Entry classification results.}
\label{tbl:results:classification}
\vspace{-3mm}
\end{table*}

\section{Experiments and Results}
\label{sec:experiments}

To conduct a set of baseline experiments on \datasetname according to the mentioned tasks, we split the data into training, validation, and test sets for all our experiments (80\%, 10\%, and 10\%, respectively). We apply stratified splitting~\cite{stratified} to maintain the same distribution of labels for each set. Implementation details of Entry Extraction (Section \ref{subsection:entryextraction}) and Entry Classification (Section \ref{subsection:entryclassification}) are available at \url{https://github.com/the-deep/humset}.

\begin{table}[h]
\centering
\begin{tabular}{l r r r}
\toprule
Model & R-1 & R-2 & R-L\\\midrule
LEAD4 & 0.32 & 0.24 & 0.31 \\
XtremeDistil$_{\mathrm{l6-h256}}$ & 0.33 & 0.25 & 0.33 \\
XLM-R$_{\mathrm{Base}}$ & \textbf{0.42} & \textbf{0.35} & \textbf{0.41}  \\
 \bottomrule
\end{tabular}
\caption{Entry extraction results for ROUGE F1 (R).}
\label{tbl:results:extraction}
\vspace{-6mm}
\end{table}

\subsection{Entry Extraction} \label{subsection:entryextraction}
\vspace{-1mm}

We evaluate the performance of the entry extraction task using ROUGE-1, ROUGE-2 ,and ROUGE-L F1 score~\cite{Lin2004}. The target text (ground truth) is a concatenation of all relevant entries, and the predicted text is a concatenation of all entries predicted as relevant. We consider a simple heuristic method (LEAD4), as well as Transformer-based~\cite{NIPS2017_3f5ee243} pre-trained language models (PLM) with a multilingual backbone as our baselines as explained in the following:

\textbf{LEAD4:} LEAD-\textit{n} is a simple baseline where the first \textit{n} sentences are predicted as being relevant entries. Consistent with prior work \cite{yela-bello-etal-2021-multihumes}, we choose $n=4$. \textbf{Transformers:} to approach the task using Transformer-based PLM, we formulate the task as a token classification problem. The objective is to distinguish between tokens that are part of relevant entries and tokens which are not. For simplicity, we fine-tune the entire model and do binary classification using a two-layer prediction head on top of the contextualized representation of each token. We conduct our experiments on XtremeDistil$_{\mathrm{l6-h256}}$~\cite{xtremedistil} and XLM-R$_{\mathrm{Base}}$~\cite{roberta} as the underlying PLM.

The evaluation results of the mentioned methods are reported in Table~\ref{tbl:results:extraction}. Among our baselines,  the model with XLM-R$_{\mathrm{Base}}$ shows the best overall performance. However, we should consider these experiments as starting points, and improvements on this task are expected by increasing model capacity and architectural variations. 

\subsection{Entry Classification}
\label{subsection:entryclassification}
\vspace{-1mm}

We test different multi-label sequence classification models applied to our five categories. We use the Precision and F1-score metrics to assess the performance of the models on each subcategory. We report macro-averages of the metrics, as the tags are unbalanced (see Table~\ref{tbl:results:classification}) and macro-averaging can provide a more nuanced view of the performance especially by supporting the more sparse classes.  Finally, we perform threshold tuning of the classification decision boundary with respect to macro-average F1-scores for each label of each category~\cite{pillai2013threshold}. Tuning the threshold is done by finding the optimal results on the validation set, used to make classifications on the test set. We conduct experiments using fastText~\cite{fasttext}, as well as Transformer-based PLMs as explained below.  

\textbf{fastText:} is an Open Source library for text representation and classification that consists of a bag of n-grams representation and a linear classifier. fastText classification is language-agnostic and does not need language-specific pre-trained word vectors, allowing us to train a multilingual classifier as a simple baseline. To handle multiple labels, we trained independent binary classifiers for each label. 
\textbf{Transformers:} For consistency with the previous task, we fine-tune the same multilingual PLMs and add a dense layer on top for multi-label classification.

Table~\ref{tbl:results:classification} reports the evaluation results on the mentioned baseline models. For comparison, a random baseline is also reported. The random baseline is a stratified random classification, created based on the distribution of the classes in the training set. Similar to the entry extraction task, the XLM-R$_{\mathrm{Base}}$ outperforms other baselines. Although overall promising results are obtained, we highlight the shortcoming of the models on the categories with many tags (Subpillars 1D and Subpillars 2D), suggesting future research directions for addressing these challenges.

\section{Conclusion}
We presented \datasetname, a new dataset of annotated humanitarian data, containing 148,621 entries with a total of 62 different tags. We have shown two NLP-based tasks that can be applied to it, providing initial experiments of its applications.
\datasetname is a multilingual human-annotated humanitarian text dataset not composed of social network data, providing a valuable and highly reliable resource for the development of automation tools regarding crisis response and humanitarian aid activities.


\section{Limitations}


\datasetname is composed of an aggregation of 46 different projects,  each with a different contribution in terms of data quantity and topics (Table \ref{table:projects}). This can introduce an implicit bias due to the different goals and themes of each project and on respective analysis framework understanding and interpretation by humanitarian annotators. \cite{DBLP:journals/corr/abs-2112-07475} refer to it as \emph{persistent subjectivity}. This is a complex and challenging limitation and, for example, \cite{geva-etal-2019-modeling} show how this kind of bias can be monitored using annotator identifiers as features in NLP models training when data is produced by crowdsourcing project \cite{crowd}. Since \datasetname is an extension of a real-case application and not the result of crowdsourcing, a more structured analysis on these aspects is needed.\\

Another complexity lies in the raw data sources. Lead text is the result of a text extraction process from PDF and HTML files (c.f. Section \ref{subsection:datasetoverview}). In both cases, converting visually-rich graphical text representation into plain text involves errors and limitations. There are several works proposing solutions for digital documents layout-aware text extraction \cite{pdf1, pdf2} but they are often domain-specific, applying only to specific types of documents. \cite{xu2020layoutlmv2} propose a Transformer-based multi-modal architecture for documents understanding using text, layout, and image data as features. Improvement in document processing could produce better data quality and subsequently improve performance on the entry extraction task (Section \ref{subsection:entryextraction}). \\

Finally, we should point out that \datasetname might contain societal biases and stereotypes and/or over-represent particular demographics or entities. This case is observed and studied in several other data resources and scenarios~\cite{bolukbasi2016man,krieg2022grep,rekabsaz_tripclick_2021}, which can lead to reflecting or even exaggerating societal biases in the system's output~\cite{melchiorre_gender_fairness,rekabsaz2020neural}, and may negatively affect users' perception and interaction behavior~\cite{krieg2022do}. Hence, when using the dataset (particularly for real-world applications), we strongly recommend first defining and monitoring such potential biases~\cite{de2019bias,rekabsaz2021measuring}, and then mitigating them using the proposed methods in literature~\cite{elazar2018adversarial,zmigrod-etal-2019-counterfactual,rekabsaz2021societal,zerveas2022mitigating,ganhoer2022mitigating}.



\section{Acknowledgement}
 We want to thank the humanitarian community users of the Data Entry and Exploration Platform (DEEP) for their openness and interest in sharing their data for this research and to trust that the NLP community can help them make their work better. We want to extend our gratitude, especially to the people working in the USAID’s Bureau for Humanitarian Assistance (BHA) for entrusting Data Friendly Space (DFS) with a grant to create the DEEP and to work on this paper; to the DEEP users and taggers for their work that make this set possible; to the project owners in DEEP that allowed us to use the data to create this set and their organizations;  to the DEEP board members Internal Displacement Monitoring Centre (IDMC), International Federation of the Red Cross, iMMAP, Office of the High Commissioner for Human Rights, Okular Analytics, United Nations Office for the Coordination of Humanitarian Affairs (UNOCHA), United Nations High Commissioner for Refugees (UNHCR), United Nations Children's FUND (UNICEF), United Nations Development Coordination Office (UNDCO), and the Danish Refugee Council (DRC). To the whole DFS team around the world, the ToggleCorp team, our partner institutions ISI Foundation and Johannes Kepler University Linz for their continuous support in making the usage of NLP possible in the humanitarian community.

\bibliography{references}
\bibliographystyle{acl_natbib}
\newpage

\appendix
\section{Additional Statistics}
\label{sec:appendix}

\begin{table}[!h]
\centering
\begin{tabular}{lrr}
\toprule
    Website &  Number &  Prop. (\%) \\
\midrule
    reliefweb.int &    5,669 &             33.6 \\
    dhakatribune.com &     834 &              4.9 \\
    redhum.org &     635 &              3.8 \\
    humanitarianresponse.info &     612 &              3.6 \\
    unb.com.bd &     380 &              2.3 \\
\midrule
    \textbf{Sum} &    8,130 &             48.2 \\
\bottomrule
\end{tabular}

\caption{The most frequently sourced websites by number and proportion of documents.}
\label{table:web-sources}
\end{table}

\begin{table}[!h]
\centering
\begin{tabular}{rrr}
\toprule
 Year &  Number &  Proportion (\textbackslash \%) \\
\midrule
 Before 2018 &      68 &              0.4 \\
 2018 &     834 &              4.9 \\
 2019 &    2,639 &             15.7 \\
 2020 &    6,087 &             36.1 \\
 2021 &    7,229 &             42.9 \\ \midrule
\textbf{Sum} &   16,857 &            100.0 \\
\bottomrule
\end{tabular}

\caption{Publishing year of documents.}
\label{table:publidh-year}

\end{table}

\begin{table}[!h]
\centering
\begin{tabular}{ L{2.5cm}  L{4cm}}
\toprule
\multicolumn{1}{c}{Region}    & Countries \\ \midrule
Africa                        &  Burkina Faso, Cameroon, Chad, Libya, Niger, Nigeria, DRC, Somalia, South Sudan, Sudan \\ \midrule
Asia \& Middle East                        &  Afghanistan, Bangladesh, Lebanon, Syria, Yemen \\ \midrule
Central/South America        & Argentina, Aruba, Bolivia, Chile, Colombia, Costa Rica, Curacao, Dominican Republic, Ecuador, El Salvador, Guatemala, Guyana, Honduras, Mexico, Panama, Paraguay, Peru, The Bahamas, Trinidad and Tobago, Uruguay, Venezuela         \\ \bottomrule
\end{tabular}
\caption{Countries of projects per region.}
\label{table:projs-per-region}
\end{table}

\begin{figure}[!h]
    \centering
  \includegraphics[width=0.5\textwidth]{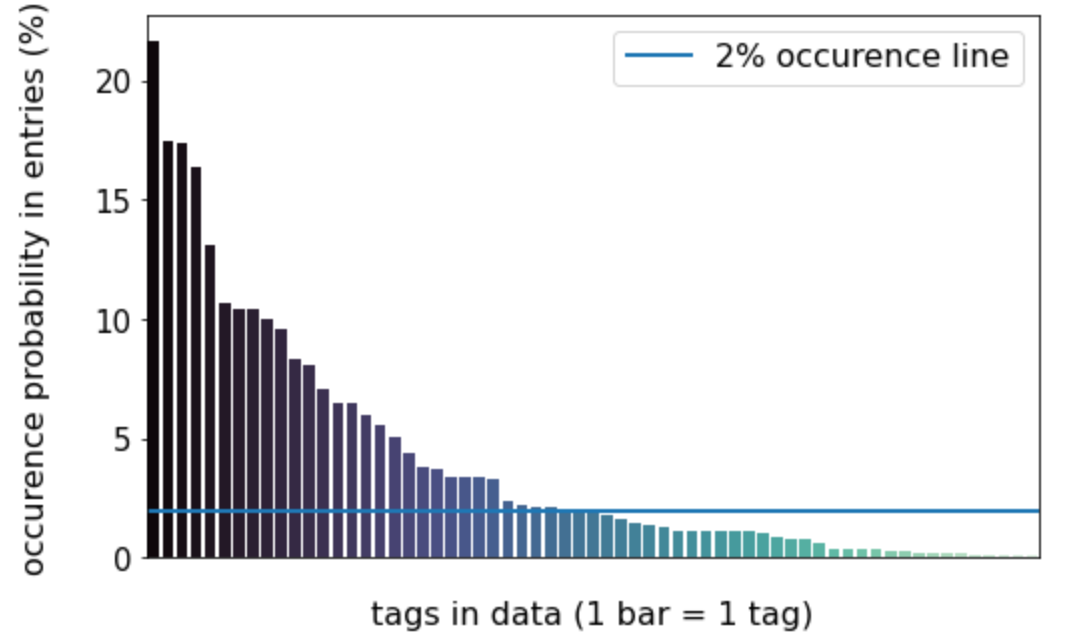}
  \caption{Proportion of tags in dataset. This figure shows the unbalanced nature of the dataset. Each bar represents a different tag. The y-axis shows the proportion of entries that contain each tag. The horizontal line added is the 2\% occurrence line. It is used to visualize the relatively high number of tags with occurrence inferior to 2\%.}
  \label{fig:histogram}
\end{figure}

\begin{table}[t]
\small
\begin{tabular}{L{4cm} r r}
\toprule
                                     Project &  \# Leads &  \# Entries\\
\midrule
                                    2020 DFS Libya &      354 &      1,581 \\
                                  2020 DFS Nigeria &      496 &      2,238 \\
COVID-19 Americas Region Multi-Sectorial Assessment &       76 &       607 \\
            Central America - Dengue Outbreak 2019 &       38 &       218 \\
          Central America: Hurricanes Eta and Iota &       48 &       363 \\
                                 GIMAC Afghanistan &      248 &      7,617 \\
                                    GIMAC Cameroon &      134 &      5,253 \\
                                        GIMAC Chad &      269 &      4,945 \\
                                       GIMAC Niger &      125 &      3,180 \\
                                     GIMAC Somalia &      157 &      5,720 \\
                                 GIMAC South Sudan &      208 &      7,212 \\
                                       GIMAC Sudan &       99 &      3,169 \\
                              IMMAP/DFS Bangladesh &     2,266 &     14,342 \\
                            IMMAP/DFS Burkina Faso &     1,279 &     13,443 \\
                                IMMAP/DFS Colombia &     1,411 &     10,760 \\
                                 IMMAP/DFS Nigeria &     1,443 &      9,620 \\
                                     IMMAP/DFS RDC &     1,586 &     13,065 \\
                                   IMMAP/DFS Syria &     1,736 &     12,043 \\
                        Lebanon Situation Analysis &       22 &       268 \\
                     Libya Situation Analysis (OA) &      681 &      2,868 \\
                   Nigeria Situation Analysis (OA) &      651 &      3,496 \\
                  Situation Analysis Generic Libya &      437 &      2,355 \\
                  Situation Analysis Generic Yemen &      371 &      2,256 \\
The Bahamas - Hurricane Dorian - Early Recovery Assessment &       17 &       191 \\
                                   UNHCR Argentina &      160 &      1430 \\
                                       UNHCR Aruba &       23 &       106 \\
                                     UNHCR Bolivia &        9 &       101 \\
                                       UNHCR Chile &      341 &      2,415 \\
                                    UNHCR Colombia &      649 &      6,349 \\
                                  UNHCR Costa Rica &       85 &       603 \\
                                     UNHCR Curacao &       20 &       111 \\
                          UNHCR Dominican Republic &       66 &       411 \\
                                     UNHCR Ecuador &      190 &      1,648 \\
                                 UNHCR El Salvador &       64 &       349 \\
                                   UNHCR Guatemala &       65 &       348 \\
                                      UNHCR Guyana &       28 &       352 \\
                                    UNHCR Honduras &       57 &       469 \\
                                      UNHCR Mexico &       16 &        96 \\
                                      UNHCR Panama &       35 &       295 \\
                                    UNHCR Paraguay &        2 &        27 \\
                                        UNHCR Peru &      247 &      2,936 \\
                         UNHCR Trinidad and Tobago &       67 &       574 \\
                                     UNHCR Uruguay &       29 &       272 \\
                                   UNHCR Venezuela &      293 &      1,805 \\
                             Venezuela crisis 2019 &      176 &       714 \\
                     Yemen Situation Analysis (OA) &       83 &       381 \\ \midrule
                     \textbf{Total}                &    16,857 &	148,602 \\
\bottomrule
\end{tabular}
\caption{Key statistics per project.}
\label{table:projects}
\end{table}

\begin{table}[!h]
\centering
\begin{tabular}{L{1.5cm} l L{4.5cm} }
\toprule
\multirow{2}{*}{Pillar~1D} & \multicolumn{2}{c}{Subpillars~1D}\\
 & \# & \multicolumn{1}{c}{Label} \\ \midrule

Context & 7 & Economy, Environment, Demography, Legal \& Policy, Politics, Security \& Stability, Sociocultural \\ \midrule
COVID-19 & 7 & Cases, Contact Tracing, Deaths, Hospitalization \& Care, Restriction Measures, Testing, Vaccination \\ \midrule
Displacement & 5 & Intentions, Local Integration, Pull Factors, Push Factors, Type / Numbers / Movements \\ \midrule
Humanitarian Access & 4 & Physical Constraints, Population to Relief, Relief to Population, Number of People Facing Humanitarian Access Constraints / Humanitarian Access Gaps \\ \midrule
Information and Communication & 4 & Communication Means and Preferences, Information Challenges and Barriers, Knowledge and Information Gaps (Hum), Knowledge And Info Gaps (Pop) \\ \midrule
Casualties & 3 & Dead, Injured, Missing \\ \midrule
Shock / Event & 3 & Hazard \& Threats, Type and Characteristics Underlying/Aggravating Factors \\

\bottomrule
\end{tabular}
\caption{Pillars and subpillars~1D in humanitarian framework.}
\label{tbl:framework:subpillar1d}
\end{table}

\begin{table}[!h]
\centering
\begin{tabular}{L{1.5cm} l L{4.5cm} }
\toprule
\multirow{2}{*}{Pillar~2D} & \multicolumn{2}{c}{Subpillars~2D}\\
 & \# & \multicolumn{1}{c}{Label} \\ \midrule
Capacities \& Response  & 4 & International Response, Local Response, National Response, Number of People Reached / Response Gaps\\ \midrule
Humanitarian Conditions  & 4 & Coping Mechanisms, Living Standards, Physical And Mental Well Being, Number of People In Need\\ \midrule
Impact &4  & Driver/Aggravating Factors, Impact on People, Impact on Systems, Services and Networks, Number of People Affected \\ \midrule
At Risk & 2 & Risk And Vulnerabilities, Number of People at Risk \\ \midrule
Priority Needs & 2 & Expressed by Humanitarian Staff, Expressed by Population \\ \midrule
Priority Interventions & 2 & Expressed by Humanitarian Staff, Expressed by Population\\ 
 \bottomrule
\end{tabular}
\caption{Pillars and subpillars~2D in humanitarian framework.}
\label{tbl:framework:subpillar2d}
\end{table}

\begin{table}[t]
\centering
\begin{tabular}{l r r }
\toprule
\multirow{2}{*}{Sectors}  & \multicolumn{2}{c}{Data Points} \\
 & Number & Proportion (\%) \\ \midrule
 Agriculture                      &      2,816      &   1.9        \\
Cross                             &      24,447     &   16.4        \\
Education                         &       9,630     &   6.5       \\ 
Food Security                     &       14,898    &   10.0         \\ 
Health                            &       32,284    &   21.7         \\ 
Livelihoods                       &       15,494    &   10.4         \\ 
Logistics                         &        2,422    &   1.6        \\ 
Nutrition                         &        5,011    &   3.4        \\ 
Protection                        &       25,986    &   17.5         \\ 
Shelter                           &       8,975     &   6.0        \\ 
WASH                              &       10,588    &   7.1         \\
\midrule
\multicolumn{1}{l}{\textbf{Count:}} & 115,176 & 77.5     \\
 \bottomrule
\end{tabular}
\caption{Proportion of sectors in the dataset.}
\label{table:sectors-nb}

\end{table}

\begin{table}[t]
\centering
\scriptsize
\begin{tabular}{L{1.7cm} L{2.2cm} r r }
\toprule
\multirow{2}{*}{Pillar~1D} & \multirow{2}{*}{Subpillar~1D} & \multicolumn{2}{c}{Data Points} \\
 &  & Number & Prop. (\%) \\ \midrule
\multirow{8}{*}{Context}                                                             &  Demography                                                                            &  3,041    &  2.0   \\ 
                                                                                    &  Economy                                                                        &  4,969   & 3.3     \\
                                                                                      & Environment                                                                       & 1,124    &  0.8   \\ 
                                                                                      & Legal \& Policy                                                                 &  2,637 & 1.8  \\ 
                                                                                      & Politics                                                                         &  1,871 & 1.3  \\ 
                                                                                      & Security \& Stability                                                            &  7,615 & 5.1 \\ 
                                                                                     &  Sociocultural                                                                     & 1,610 & 1.1 \\\cdashlinelr{2-4}   
 & \multicolumn{1}{l}{\textbf{Pillar~1D Count:}} & 21,070 & 14.2 \\ \midrule 
\multirow{8}{*}{COVID-19}                                                            & Cases                                                                               & 5,501 & 3.7 \\
                                                                                     & Contact Tracing                                                                     & 627 & 0.4 \\ 
                                                                                     & Deaths                                                                              & 3,095 & 2.1 \\ 
                                                                                     & Hospitalization \& Care                                                              & 510 & 0.3 \\
                                                                                     & Restriction Measures                                                                & 5,006 & 3.4 \\ 
                                                                                     & Testing                                                                             & 1,621 & 1.1 \\ 
                                                                                     & Vaccination                                                                         & 2,764 & 1.9 \\\cdashlinelr{2-4}  
 & \multicolumn{1}{l}{\textbf{Pillar~1D Count:}} & 14,964 & 10.1 \\ \midrule 
\multirow{6}{*}{Displacement}                                                        & Intentions                                                                         & 454 & 0.3 \\ 
                                                                                     & Local Integration                                                                   & 1,657 & 1.1 \\ 
                                                                                     & Pull Factors                                                                       & 342 & 0.2 \\ 
                                                                                     & Push Factors                                                                       & 2,047 & 1.4 \\ 
                                                                                     &    Type / Numbers / Movements                                                          & 8,280 & 5.6 \\\cdashlinelr{2-4}  
 & \multicolumn{1}{l}{\textbf{Pillar~1D Count:}} & 10,678 & 7.2 \\ \midrule 
\multirow{5}{*}{\begin{tabular}[c]{@{}l@{}}Humanitarian \\ Access\end{tabular}}      & Number Of People Facing Humanitarian Access Constraints/Humanitarian Access Gaps & 658 & 0.4 \\ 
                                                                                     &  Physical Constraints                                                            & 1,483 & 1.0 \\  
                                                                                     &  Population to Relief                                                            & 204 & 0.1 \\  
                                                                                     & Relief to Population                                                             & 863 & 0.6 \\\cdashlinelr{2-4}   
 & \multicolumn{1}{l}{\textbf{Pillar~1D Count:}} & 2,922 & 2.0 \\ \midrule 
\multirow{5}{*}{\begin{tabular}[c]{@{}l@{}}Information\\and\\Communication\end{tabular}} & Communication Means and Preferences                                          & 150 & 0.1 \\ 
                                                                                     & Information Challenges and Barriers                                              & 100 & 0.1 \\  
                                                                                     & Knowledge and Info Gaps (Hum)                                                    & 573 & 0.4 \\  
                                                                                     & Knowledge and Info Gaps (Pop)                                                    & 218 & 0.1 \\  \cdashlinelr{2-4} 
 & \multicolumn{1}{l}{\textbf{Pillar~1D Count:}} & 993 & 0.7 \\ \midrule 
\multirow{4}{*}{Casualties}                                                          & Dead                                                                            &  3,147   &  2.1   \\ 
                                                                                     & Injured                                                                         &   649     &  0.4 \\  
                                                                                     & Missing                                                                         & 269        &  0.4 \\  \cdashlinelr{2-4} 
 & \multicolumn{1}{l}{\textbf{Pillar~1D Count:}} & 3,497 & 2.4 \\ \midrule
 \multirow{4}{*}{Shock/Event}                                                         & Hazard \& Threats                                                               & 3,616 & 2.4 \\ 
                                                                                     & Type and Characteristics                                                        & 1,241 & 0.8\\  
                                                                                     & Underlying/Aggravating Factors                                                  & 1,589 & 1.1 \\  \cdashlinelr{2-4} 
 & \multicolumn{1}{l}{\textbf{Pillar~1D Count:}} & 6,072 & 4.1 \\ \cdashlinelr{1-4} 
  
 & \multicolumn{1}{l}{\textbf{Overall Count:}} & 53,575 & 36.0 \\ 
 \bottomrule
\end{tabular}
\caption{Proportion of each pillar and subpillar~1D in the dataset.}
\label{table:subpillars1d-nb}

\end{table}

\begin{table}[h]
\centering
\small
\begin{tabular}{L{1.4cm} L{2.6cm} r r }
\toprule
\multirow{2}{*}{Pillar~2D} & \multirow{2}{*}{Subpillars~2D} & \multicolumn{2}{c}{Data Points} \\
 &  & Number & Prop. (\%) \\ \midrule
\multirow{5}{*}{\begin{tabular}[c]{@{}l@{}}Capacities \& \\ Response\end{tabular}}   & International Response                                                              & 15,422 & 10.4 \\ 
                                                                                     &  Local Response                                                                      & 211 & 0.1 \\
                                                                                     & National Response                                                                   & 6,596 & 4.4 \\ 
                                                                                      &  Number Of People Reached / Response Gaps                                         & 3,316 & 2.2 \\\cdashlinelr{2-4}   
 & \multicolumn{1}{l}{\textbf{Pillar~2D Count:}} & 20,272 & 13.6 \\ \midrule 
\multirow{5}{*}{\begin{tabular}[c]{@{}l@{}}Humanitarian \\ Conditions\end{tabular}}  & Coping Mechanisms                                                                   & 4,994 & 3.4 \\
                                                                                     & Living Standards                                                                    & 25,922 & 17.4 \\ 
                                                                                     & Number of People In Need                                                            & 1,315 & 0.9 \\
                                                                                    & Physical and Mental Well Being                                                      & 15,845 & 10.7 \\\cdashlinelr{2-4}  
 & \multicolumn{1}{l}{\textbf{Pillar~2D Count:}} & 43,301 & 29.1 \\ \midrule 
\multirow{5}{*}{Impact}                                                              & Driver/Aggravating Factors                                                          & 12,029 & 8.1 \\ 
                                                                                     &    Impact on People                                                                 & 12,319 & 8.3 \\ 
                                                                                     & Impact on Systems, Services And Networks                                            & 14,335 & 9.6 \\  
                                                                                     &  Number of People Affected                                                          & 2,293 & 1.5 \\\cdashlinelr{2-4}  
 & \multicolumn{1}{l}{\textbf{Pillar~2D Count:}} & 33,699 & 22.7 \\ \midrule 
\multirow{3}{*}{At Risk}                                                             & Risk and Vulnerabilities                                                            & 9,625 & 6.5 \\ 
                                                                                     & Number of People Aa Risk                                                            & 261 & 0.2 \\\cdashlinelr{2-4}   
 & \multicolumn{1}{l}{\textbf{Pillar~2D Count:}} & 9,625 & 6.5 \\ \midrule 
\multirow{3}{*}{Priority Needs}                                                      & Expressed by Humanitarian Staff                                                     & 1,664 & 1.1 \\ 
                                                                                     & Expressed by Population                                                             & 1,669 & 1.1 \\\cdashlinelr{2-4}   
 & \multicolumn{1}{l}{\textbf{Pillar~2D Count:}} & 3,272 & 2.2 \\ \midrule 
\multirow{2}{*}{\begin{tabular}[c]{@{}l@{}}Priority \\ Interventions\end{tabular}}   & Expressed by Humanitarian Staff                                                     & 5,575 & 3.8 \\ 
                                                                                     & Expressed by Population                                                             & 287 & 0.2 \\ \cdashlinelr{2-4}  
 & \multicolumn{1}{l}{\textbf{Pillar~2D Count:}} & 5,844 & 3.9 \\ \cdashlinelr{1-4} 
  
 & \multicolumn{1}{l}{\textbf{Overall Count:}} & 94,429 & 63.5 \\ 
 \bottomrule
\end{tabular}
\caption{Proportion of each pillar and subpillar~2D in the dataset.}
\label{table:subpillars2d-nb}

\end{table}

\end{document}